\documentclass[12pt,a4paper]{article}
\usepackage{float}
\usepackage{rotating}
\usepackage{natbib}
\usepackage{xcolor}
\usepackage{soul}

\usepackage{libertinus}
\usepackage[a4paper,vmargin={20mm,20mm},hmargin={20mm,20mm}]{geometry}
\usepackage{amsmath, amssymb, amsfonts, amsthm, mathrsfs}

\usepackage{array, multirow, tabularx, booktabs, longtable, colortbl, adjustbox}
\usepackage{hyperref}
\usepackage{graphicx, subcaption, float, rotating, placeins}

\usepackage{caption, xcolor, textcomp, url, enumitem, scrextend, blindtext}

\usepackage{algorithm, algorithmicx, algpseudocode}
\usepackage{siunitx}
\usepackage{authblk}        
\usepackage{geometry}   
\usepackage{subcaption}
\usepackage[title]{appendix} 
\usepackage{manyfoot} 
\usepackage[table]{xcolor}
\usepackage{colortbl}
\usepackage{adjustbox}
\theoremstyle{definition} 
\newtheorem{definition}{Definition}

\emergencystretch=\maxdimen
\begin{document}
\title{Why Neural Networks Forget: Catastrophic Forgetting Through the Lens of Sloppiness}
\date{}
\author[1]{Hari Priya Narahari}
\author[1]{Karthik Raman}
\author[2]{Prem Jagadeesan\thanks{Corresponding author: \texttt{j\_prem@cb.amrita.edu}}}


\affil[1]{Department of Data Science and AI, Wadhwani School of Data Science and AI, IIT Madras, Chennai, Tamil Nadu, India}
\affil[2]{Amrita School of Artificial Intelligence Coimbatore, Amrita Vishwa Vidyapeetham, Tamil Nadu, India}

\maketitle
\begin{abstract}
Catastrophic forgetting has been one of the biggest challenges of continual learning in Neural Networks (NNs). Several studies have formulated multiple techniques and methods to circumvent this problem. However, an optimization theory-based framework of forgetting using geometry is less explored. Using the geometry of parameter spaces, we present a framework through sloppiness to explain catastrophic forgetting and a solution to circumvent it in continual learning. We show that for continual learning,  training networks along sloppy directions can minimise the amount of forgetting. Our method performs better than current projection and first-order methods, particularly in situations where forgetting happens frequently.
\end{abstract}

\section{Introduction}
Neural Networks (NNs) have existed for several decades. It is not surprising that these seemingly complex networks lie at the heart of all advances and learning in AI, given their capabilities. NNs are models that are capable of decoding underlying patterns in large datasets. The initial source of inspiration for the design was biological neurons and how our brains make decisions from data \cite{mcculloch1943logical, rosenblatt1958perceptron, lecun2015deep}.\\\\The foundational work of McCulloch and Pitts \cite{mcculloch1943logical} considers the neuron as binary logical gates, while modern Neural Networks are more complex in terms of their underlying architecture involving backpropagation, introduced by Rumelhart et al. for iteratively updating and arriving at the optimal parameter set for the network \cite{rumelhart1986learning}.\\\\These extremely useful models are no exception to having a few disadvantages, particularly the ability to retain previously acquired knowledge when learning sequentially. The unexpected unlearning that happens when NNs are trained on conflicting tasks sequentially is called catastrophic forgetting \cite{french1999catastrophic, goodfellow2013empirical}. McCloskey et al. \cite{mccloskey1989catastrophic} describe catastrophic forgetting as the process of losing previously acquired knowledge when learning a new task. Training on a sequence of different tasks results in accuracy on the first task dropping back to chance levels as subsequent tasks are learned \cite{kirkpatrick2017overcoming, zenke2017continual}. A neural network's configuration can be represented by a point in a multi-dimensional weight space. When a network is trained for multiple tasks, it is expected that this configuration lies at the intersection of the solution spaces of the tasks it has been trained for.\\\\Learning a task, say task $t_1$, is minimizing the loss for task $t_1$, defined as $\mathcal{L_{\text{$t_1$}}(\theta)}$. $\theta^*_{t_1}$ denotes the optimum for task $t_1$ where $\left.\nabla_{\theta}\mathcal{L}_{t_1}\right|_{\theta = \theta^*} = 0$. Training the same network on task $t_2$ sequentially, minimizes the loss $\mathcal{L}_{t_2}(\theta)$. Now, if gradients that minimize both loss functions don't align with each other, then gradient updates that minimize $\mathcal{L}_{t_2}(\theta)$ can increase $\mathcal{L}_{t_1}(\theta)$, resulting in performance degradation on task $t_1$. It is important to note that the effect of catastrophic forgetting is severe when there are conflicting tasks, leading to conflicting gradient directions in parameter space.\\\\ Researchers have dedicated significant effort to formulating methods and frameworks to overcome catastrophic forgetting. Some noteworthy work includes the seminal work by Kirkpatrick et al.  \cite{kirkpatrick2017overcoming}, which proposes Elastic Weight Consolidation (EWC) as a solution to catastrophic forgetting and demonstrates continual learning. The work assessed parameter importance through the Bayesian framework by finding the posterior probability for a set of parameters in a given data (training) distribution. Using the diagonal of the Fisher Information Matrix, the curvature of the loss function near the optimum is measured. High values of FIM indicate significant changes in the loss for small parameter changes, while low values have a much less impact on the loss. To prevent forgetting, a quadratic penalty is added to the loss function to penalize changes to the most important parameters. Thus, this allows non-important parameters to change freely while learning the new task.
Synaptic Intelligence (SI) \cite{zenke2017continual}, by Zenke et al., proposes intelligent synapses that enable the network to remember by storing additional information. The architecture is designed not only to retain current and past weights after training a new task, but also to capture the importance of those weights, making it a three-dimensional state space. A penalty is added if important weights are altered. The importance is calculated as the network is trained, making it dynamic.
Functionally invariant paths (FIPs) by Raghavan et al. \cite{raghavan2024engineering} exploit parameter degeneracy, which governs over-parameterized models in which many different weight configurations can achieve the same performance on a given task.
Unlike traditional methods that treat a trained network as a single, fixed point in weight space, FIP conceptualizes the weight space as a curved Riemannian manifold. Ref.~\cite{kemker2018measuring} explores new benchmarks and metrics to overcome catastrophic forgetting.
Current projection-based continual learning methods update their parameters orthogonally in relation to first-order gradient spaces. Notable examples include Orthogonal Gradient Descent (OGD) \cite{farajtabar2020orthogonal} that restricts updates to be orthogonal to the first-order gradient, Gradient Projection Memory (GPM) \cite{saha2021gradient} which identifies core task subspaces within the first-order gradient space and constrains updates to their orthogonal complement using Singular Value Decomposition (SVD) and orthogonal weight modification (OWM) \cite{zeng2019continual}, which limits parameter adjustments to reside within the null space of input feature covariances.\\\\ In the current work, we establish a geometric reasoning that identifies the cause of catastrophic forgetting using the sloppiness framework. Here, we demonstrate that "parameter updates in certain (stiff) directions are indeed responsible for the forgetting". By constraining a network to only the stiff directions, we observe an accuracy drop similar to that in an unconstrained network, establishing these directions as the primary drivers of catastrophic forgetting in a task-conflicting continual learning setting. Our framework seeks to address these challenges through the application of second-order geometry. By framing catastrophic forgetting through the lens of sloppiness, we show that it is not merely gradient alignment, but specifically updates along high-curvature (stiff) Hessian eigenspaces that drive forgetting. This provides a principled, curvature-based justification for orthogonal projection in parameter space. To summarize our contributions:
\begin{itemize}
\item We put forth a geometric explanation for catastrophic forgetting based on the idea of sloppiness in Neural Networks.
\item We hypothesize that catastrophic forgetting primarily arises from updates along stiff directions of the loss landscape of a trained network.
To demonstrate this, we introduce a constrained optimization procedure that isolates updates along stiff and sloppy subspaces during training in a task-conflicting learning setting.
\item We demonstrate that restricting updates to stiff directions reproduces catastrophic forgetting behaviour, while updates along sloppy directions preserve previous task performance.
\end{itemize}
 \begin{figure} [htbp]
     \centering
     \includegraphics[width=1\linewidth]{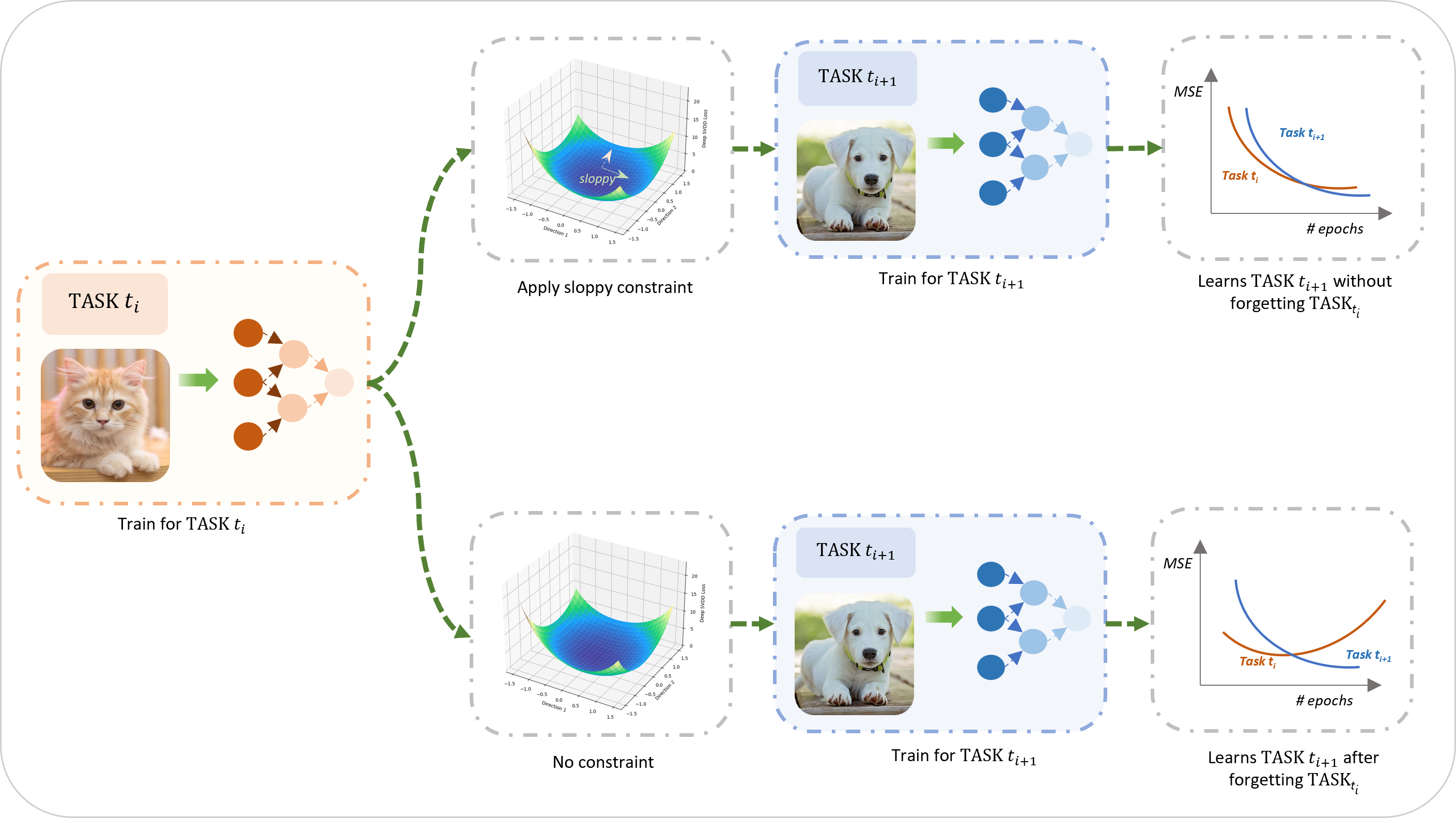}
     \caption{An illustration of the constrained optimization procedure that isolates stiff directions and updates along sloppy subspaces
during training in a task-conflicting learning setting}
     \label{fig:sloppy-fig}
 \end{figure}
\section{Background}
Sloppiness is the network's ability to tolerate significant changes in parameters while still maintaining functional accuracy for a given task. It is said that these parameter spaces where predictions are identical are abundant, due to over-parameterisation in NNs~\cite{mao2026analytical}.
Existing literature quantifies sloppiness\cite{jagadeesan2023sloppiness} as
\begin{equation}
     S = \frac{\lambda_{\text{min}}(H)}{\lambda_{\text{max}}(H)}
\end{equation}
where $H$ is the Hessian of the loss function and $H = \nabla^2_\theta\mathcal{L}(\theta)$. The $\lambda_{\text{min}}(H)$ corresponds to the eigenvalue of the most sloppy direction, and $\lambda_{\text{max}}(H)$ is the eigenvalue of the stiffest direction of the given model. In other words, a model is considered sloppy when the eigenvalues of the Hessian span many orders of magnitude, resulting in a hierarchy of stiff and sloppy parameter directions\cite{gutenkunst2007universally}. The main challenge is identifying all sloppy regions in a non-linear parameter space, such as that of a neural network, which is usually infeasible. It can be an indicator of the presence of regions in the parameter space where the predictions of the model are nearly identical \cite{gutenkunst2007universally}. Hence, sloppiness is defined locally for a given parameter.\\\\
Sloppiness becomes a crucial aspect in data-driven modeling, as it is closely connected to parameter identifiability. Sloppy regions are the main cause of large uncertainties in parameter estimates. The loss landscape of a neural network is usually non-convex and changes more rapidly in some regions than others. Thus, second-order geometry is better because it measures how sensitive the loss is to changes in parameters, not just which directions have been used before.
\begin{figure} [htbp]
     \centering
     \includegraphics[width=0.75\linewidth]{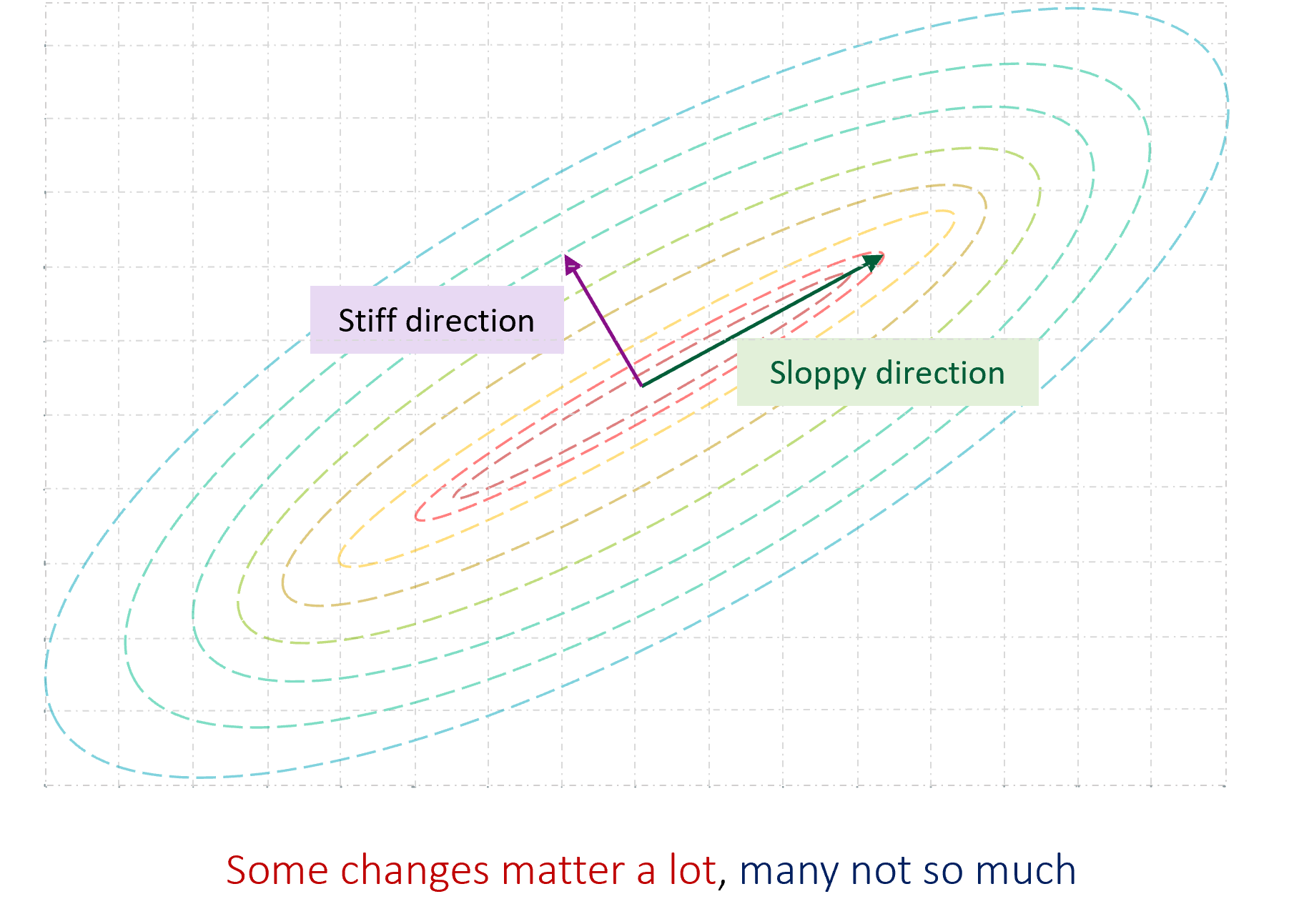}
     \caption{\textbf{Sloppiness illustrated}\\Sloppy directions are those characterised by eigenvectors with small curvature, by which the loss landscape does not change much. Stiff directions are those characterised by eigenvectors with large curvature whose perturbations lead to a drastic change in the loss landscape.} 
     \label{fig:sloppy-fig}
 \end{figure}

\section{Methods}
To establish a framework for catastrophic forgetting using the notion of sloppiness, we hypothesize that forgetting occurs when updates during network learning are made along stiff directions in parameter space. Stiff directions are defined locally around an optimal parameter configuration as the eigenvectors corresponding to the top $k$ largest eigenvalues of the cost function's Hessian. Updates in these directions induce significant changes in the loss and can therefore strongly interfere with previously learned tasks, leading to catastrophic forgetting. \\\\
Using this, we aim to demonstrate the stiff directions essential for forgetting when they are altered during sequential learning of a task. For this, we opt to constrain the parameter space to learn through three modes: (1) sloppy mode, (2) stiff mode, and (3) none
\\\\ Sloppy mode allows the parameters to be altered only in the sloppy direction. Stiff mode constrains the network to alter only the stiff directions during learning, whereas the none mode imposes no constraints.
By showing forgetting in the (2) stiff constraint, which happens similarly to the (3) none mode in a task-conflicting setting, we can demonstrate that the parameter updates in stiff directions are indeed responsible for forgetting.\\\\
We train a Deep SVDD (Support Vector Data Description)\cite{ruff2018deep} network to learn (four) tasks sequentially: Task $t_i$ followed by Task $t_{i+1}$. When the network learns Task $t_i$, a specific center $c_i$ is chosen that it learns to map all images of one digit (e.g., 1). For Task $t_{i+1}$, we select a new center $c_{i+1}$, orthogonal to the previous center, ensuring their dot product is zero, and train the network to map images of a different digit (e.g., 8) to $c_{i+1}$. After training Task $t_{i+1}$, we demonstrate that the network fails to map digit 1 to $c_i$ accurately. The accuracy drop demonstrates catastrophic forgetting.\\\\
The Deep SVDD\cite{ruff2018deep} architecture takes in a 784-dimensional input vector, e.g., flattened MNIST images\cite{lecun1998gradient} and passes it through two hidden layers (10 neurons, then 8 neurons) with ReLU activations. The output is a $k$-dimensional vector where $k=4$. The initial network is trained with the Adam\cite{kingma2014adam} optimizer. The loss function for the network is given by
\begin{equation}
     \mathcal{L}_{\text{SVDD}} = \frac{1}{N}\sum_{i=1}^{N} || f_\theta(x_i) - c||^2
\end{equation}
Accuracy is measured by the distance $d$ between the target center $c$ and the mapped digit $f_\theta(x_i)$. A smaller distance is a measure of higher accuracy.
\begin{equation}
     d= || f_\theta(x_i) - c||
\end{equation}
We also train a multi-head LeNet\cite{lecun1998gradient} architecture to understand how our method works with a classification task. The LeNet architecture consists of
\begin{itemize}
    \item 2 convolutional layers, each followed by a ReLU activation function and a max-pooling layer
    \begin{itemize}
        \item $1^{st}$ convolutional layer (1 $\to$ 6, kernel size: 5), ReLU activation, Max pooling (2, 2)
        \item $2^{nd}$ convolutional layer (6 $\to$ 16, kernel size: 5), ReLU activation, Max pooling (2, 2)
     \end{itemize}
    \item 2 Fully Connected Layers 
    \begin{itemize}
        \item $1^{st}$ fully connected layer (16 * 5 * 5 $\to$ 120), ReLU activation
        \item $2^{nd}$ fully connected layer (120 $\to$ 84), ReLU activation
    \end{itemize}
The loss function for the network is given by
\begin{equation}
     \mathcal{L}_{\text{LeNet}} = -\sum_{i=1}^{N}y_i\log(p(y_i)) - (1-y_i)\log(1-p(y_i)) 
\end{equation}
\end{itemize}
Performance is measured by the accuracy which is the ratio of correct predictions out of all predictions made.
To demonstrate that updates made in stiff directions drive catastrophic forgetting in a task-conflicting setting, we train Task $t_{i+1}$ under three constraint modes:
\begin{itemize}
    \item \textbf{Sloppy-constrained optimization (Sloppy mode)}: This ensures the parameters are only modified along the sloppy directions of the Hessian of Task $t_i$.
    \item \textbf{Stiff-constrained optimization (Stiff mode)}: This ensures the parameters are only modified along the stiff directions of the Hessian of Task $t_i$.
    \item \textbf{Unconstrained optimization (None mode)}: No constraint is applied, and the network is trained as usual.
\end{itemize} 

\begin{definition}[Sloppy-constrained subspace] 
The sloppy-constrained subspace can be defined as
\label{thm:sloppy-subspace}
\begin{equation}
\Delta_{t_{i+1}}^{(\mathrm{sloppy})}
=
\left(I-P_{t_{i}}^{(\mathrm{stiff})}\right)\Delta_{t_{i+1}}.
\end{equation}
\end{definition}
\begin{proof}
The projection matrix onto the stiff subspace of Task $t_i$ is
\begin{equation}
P_{t_{i}}^{(\mathrm{stiff})}
=
\sum_{i=1}^{k} v_i^{({t_{i}})}v_i^{({t_{i}})\top}
=
\mathrm{Basis}_{t_{i}}\mathrm{Basis}_{t_{i}}^\top,
\end{equation}
where
\begin{equation}
\mathrm{Basis}_{t_{i}}
=
\begin{bmatrix}
v_1^{({t_{i}})} & v_2^{({t_{i}})} & \cdots & v_k^{({t_{i}})}
\end{bmatrix},
\qquad
\mathrm{Basis}_{t_{i}}^\top \mathrm{Basis}_{t_{i}} = I_k.
\end{equation}

Let
\begin{equation}
\Delta_{t_{i+1}} = \theta_{t_{i+1}} - \theta_{t_{i}}^*
\end{equation}
denote the parameter update after training on Task ${t_{i+1}}$.

The projection of $\Delta_{t_{i+1}}$ onto the stiff subspace is
\begin{equation}
\Delta_{t_{i+1}}^{(\mathrm{stiff})}
=
P_{t_{i}}^{(\mathrm{stiff})}\Delta_{t_{i+1}}.
\end{equation}

Since every vector can be uniquely decomposed into components lying in a subspace and its orthogonal complement,
\begin{equation}
\Delta_{t_{i+1}}
=
\Delta_{t_{i+1}}^{(\mathrm{stiff})}
+
\Delta_{t_{i+1}}^{(\mathrm{sloppy})}.
\end{equation}

Therefore,
\begin{align}
\Delta_{t_{i+1}}^{(\mathrm{sloppy})}
&=
\Delta_{t_{i+1}}-\Delta_{t_{i+1}}^{(\mathrm{stiff})} \\
&=
\Delta_{t_{i+1}}-P_{t_{i}}^{(\mathrm{stiff})}\Delta_{t_{i+1}} \\
&=
\left(I-P_{t_{i}}^{(\mathrm{stiff})}\right)\Delta_{t_{i+1}}.
\end{align}

Hence,
\begin{equation}
\Delta_{t_{i+1}}^{(\mathrm{sloppy})}
=
\left(I-P_{t_{i}}^{(\mathrm{stiff})}\right)\Delta_{t_{i+1}}.
\end{equation}
\end{proof}

\begin{definition}[Gradient update for sloppy constrained optimization] \label{thm:grad-update}
The gradient update for the sloppy subspace is given by:
\begin{equation}
\theta_{t_{i+1}}^{(\tau+1)}
=
\theta_{t_{i+1}}^{(\tau)}
-
\eta
\left(I-P_{t_{i}}^{(\mathrm{stiff})}\right)
\nabla_\theta
\mathcal L_{t_{i+1}}(\theta_{t_{i+1}}^{(\tau)}).
\end{equation}
\end{definition}

\begin{proof}
By definition \ref{thm:sloppy-subspace}, the projection onto the sloppy subspace is
\begin{equation}
P_{t_{i}}^{(\mathrm{sloppy})}
=
I-P_{t_{i}}^{(\mathrm{stiff})}.
\end{equation}

At iteration $\tau$, the standard gradient descent update for Task ${t_{i+1}}$ is
\begin{equation}
\theta_{t_{i+1}}^{(\tau+1)}
=
\theta_{t_{i+1}}^{(\tau)}
-
\eta
\nabla_\theta \mathcal L_{t_{i+1}}(\theta_{t_{i+1}}^{(\tau)}).
\end{equation}

To constrain the optimization to the sloppy subspace, the gradient is first projected using
$P_{t_{i}}^{(\mathrm{sloppy})}$. Hence,
\begin{equation}
\nabla_\theta \mathcal L_{t_{i+1}}(\theta_{t_{i+1}}^{(\tau)})
\longrightarrow
P_{t_{i}}^{(\mathrm{sloppy})}
\nabla_\theta \mathcal L_{t_{i+1}}(\theta_{t_{i+1}}^{(\tau)}).
\end{equation}

Substituting
$
P_{t_{i}}^{(\mathrm{sloppy})}
=
I-P_{t_{i}}^{(\mathrm{stiff})}
$
into the gradient descent update yields
\begin{equation}
\theta_{t_{i+1}}^{(\tau+1)}
=
\theta_{t_{i+1}}^{(\tau)}
-
\eta
\left(I-P_{t_{i}}^{(\mathrm{stiff})}\right)
\nabla_\theta
\mathcal L_{t_{i+1}}(\theta_{t_{i+1}}^{(\tau)}).
\end{equation}
Thus, each update is restricted to the sloppy subspace.
\end{proof}

For a given network, after training on Task $t_i$ at optimum $\theta_{t_i}^*$, we identify the top $k$ eigenvalues and corresponding eigenvectors of the Hessian ($H$) of the cost function. These correspond to the stiff directions. The stiff subspace is given by the span of these top $k$ eigenvectors projected from $\theta^*$ of Task $t_i$. The sloppy subspace comprises the orthogonal complement of the stiff subspace, i.e., all remaining parameter directions. The subspace is computed each time before the network is trained on the new task $t_{i+1}$ using the best model obtained from the previous task.\\

\begin{algorithm}[htbp]
    \caption{Quantifying Catastrophic Forgetting}
    \label{alg:cf}
    \begin{algorithmic}[1]
        \Statex \textbf{Aim:} To demonstrate that learning in stiff directions results in catastrophic forgetting 
        \Require Tasks, $\text{Task } t_1$, $t_2$, ..., $\text{Task } t_n$
        \For{each task $t_i \in \mathcal{T}$}
            \State Train a neural network on Task ${t_{i}}$
            \State Obtain the optimal parameters $\theta_{t_{i}}^{*}$
            \State Compute the Hessian of the loss function for all tasks up to Task ${t_{i}}$:
            \[H_{t_{i}} = \nabla_\theta^2 \mathcal{L}_{t_{i}}(\theta) |_{\theta=\theta_{t_{i}}^*}\]
            \State Extract the top $k$ eigenvalues and corresponding eigenvectors of $H_{t_{i}}$:
            \[
                \{ (\lambda_i^{({t_{i}})}, v_i^{({t_{i}})}) \}_{i=1}^{k}
            \]
            
            Train Task ${t_{i+1}}$ starting from $\theta_{t_{i}}^{*}$ under two conditions:
            \begin{enumerate}
                \item Stiff-constrained optimization mode (stiff)
                \item Unconstrained optimization mode (none)
            \end{enumerate}
              \State Project gradients onto stiff subspace $P_{t_{i}}^{\text{(stiff)}}$ if stiff mode
            
            \State \textbf{return} trained parameters 
            $\{ \theta_{t_{i+1}}^{(\text{stiff})}, \, \theta_{t_{i+1}}^{(\text{none})} \}$ to compare retention levels of Task ${t_{i}}$ and Task ${t_{i+1}}$ (stiff and none modes) by the network.
        \EndFor
    \end{algorithmic}
\end{algorithm}
\begin{algorithm}[htbp]
    \caption{Tackling Catastrophic Forgetting through Sloppiness}
    \label{alg:sloppy_cf}
    \begin{algorithmic}[1]
        \Require Tasks, $\text{Task } t_1$, $t_2$, ..., $\text{Task } t_n$
        \For{each task $t_i \in \mathcal{T}$}
            \State Train a neural network on Task ${t_{i}}$
            \State Obtain the optimal parameters $\theta_{t_{i}}^{*}$
            \State Compute the Hessian of the loss function for the model trained up to Task ${t_{i}}$:
            \[H_{t_{i}} = \nabla_\theta^2 \mathcal{L}_{t_{i}}(\theta) |_{\theta=\theta_{t_{i}}^*}\]
            \State Extract the top $k$ eigenvalues and corresponding eigenvectors of $H_{t_{i}}$:
            \[
                \{ (\lambda_i^{({t_{i}})}, v_i^{({t_{i}})}) \}_{i=1}^{k}
            \]
            \State Train on Task ${t_{i+1}}$ starting from $\theta_{t_{i}}^{*}$ in the sloppy-constrained mode by projecting gradients onto sloppy subspace
        \EndFor
        \State \textbf{return} trained parameters 
    \end{algorithmic}
\end{algorithm}

\begin{figure} [htb]
\centering     
\includegraphics[width=0.75\linewidth]{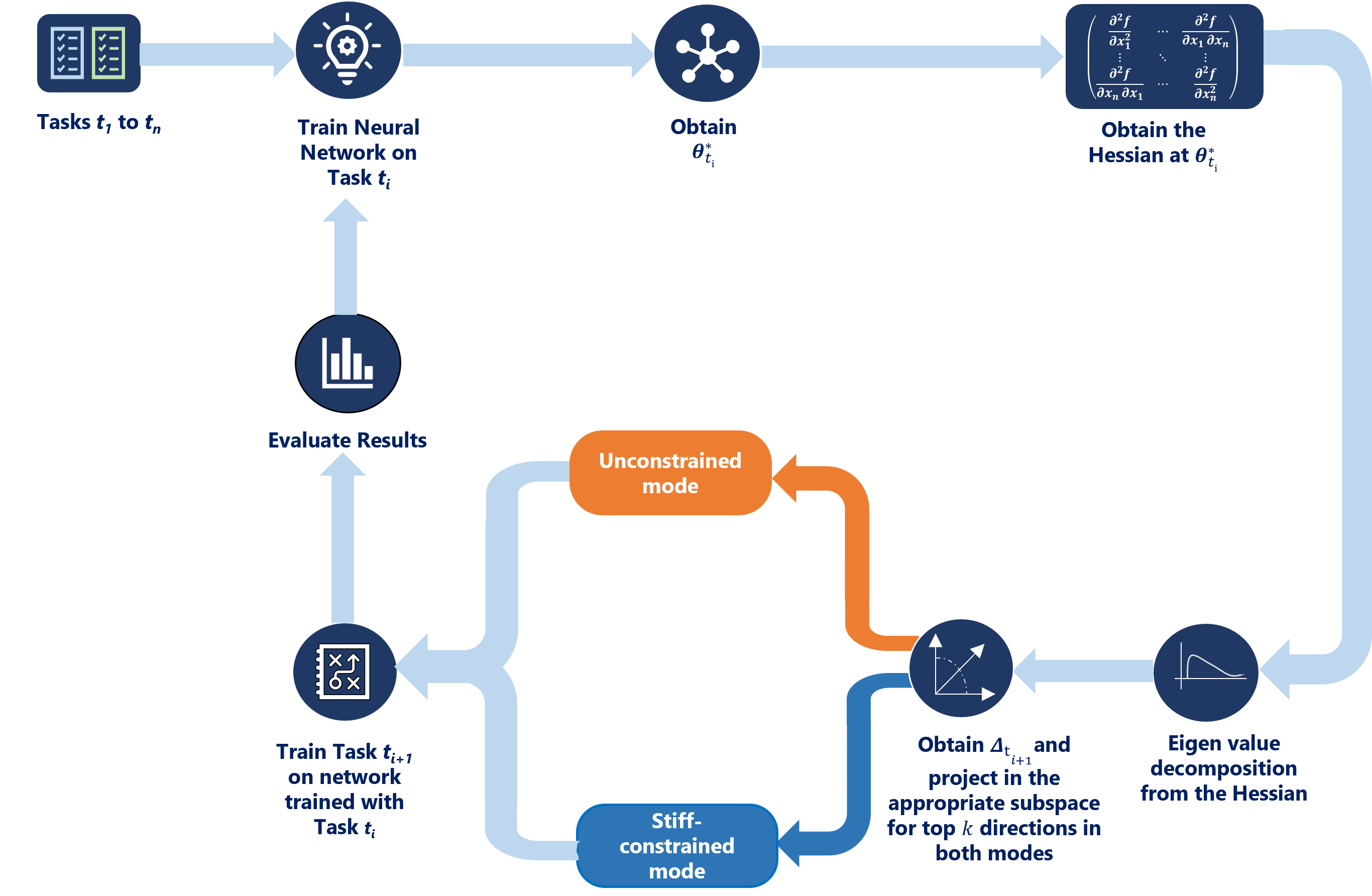}
     \caption{\textbf{Algorithm 1 for tasks illustrated}\\We start with $n$ task-conflicting tasks, Task $t_1$ and Task $t_n$. A Neural Network is initially trained on Task $t_i$, producing optimum $\theta^*_{t_i}$. The Hessian at $\theta^*_{t_i}$ is used to obtain the top $k$ eigen directions. Next, Task $t_{i+1}$ is trained on the Neural Network that was initially trained for Task $t_i$, using two different constraint modes. The results from these modes are evaluated, which will demonstrate that stiff directions are indeed responsible for forgetting. The process is repeated until all tasks are trained. }
     \label{fig:fig2}
 \end{figure}
\begin{figure} [htb]
     \centering
     \includegraphics[width=0.8\linewidth]{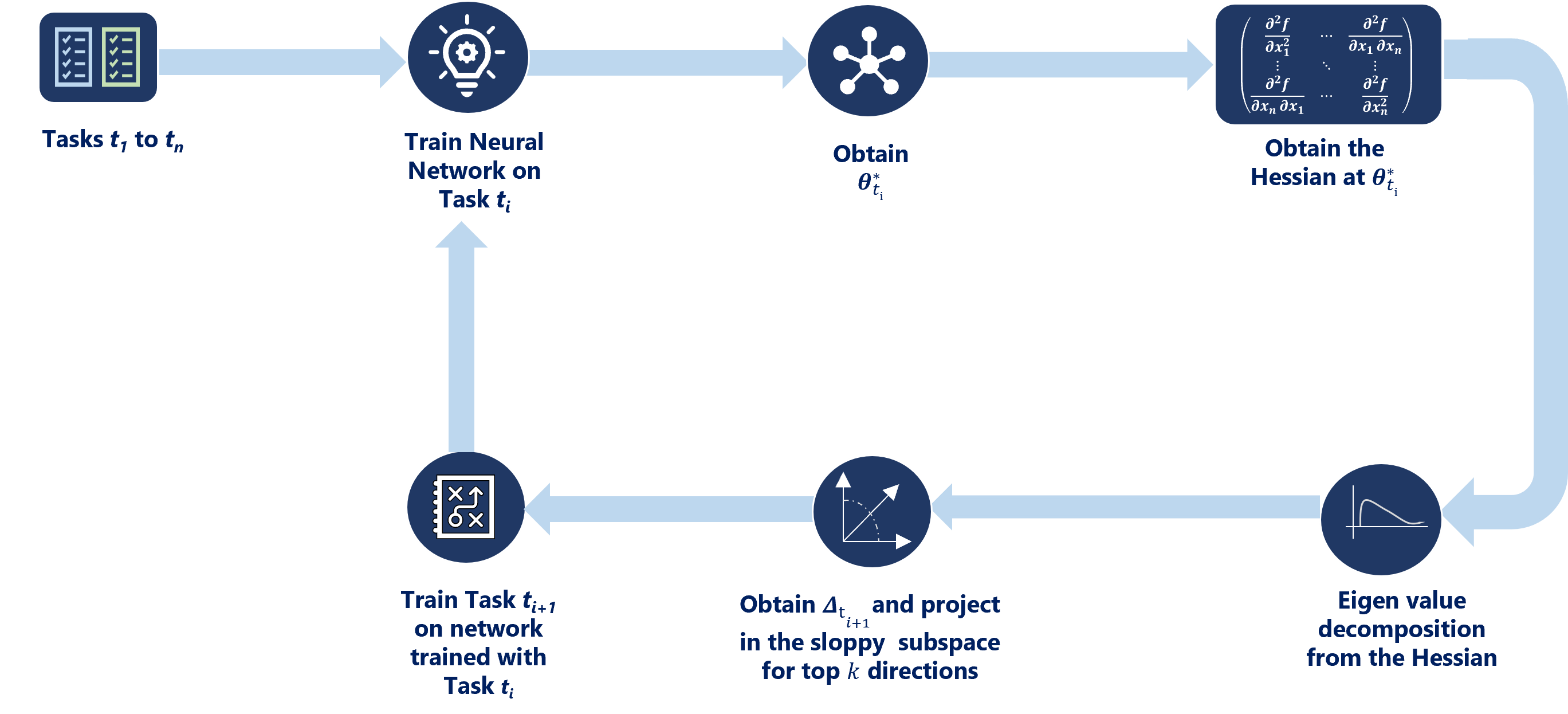}
     \caption{\textbf{Algorithm 2 illustrated}: We start with $n$  tasks, Task $1$ and Task $n$. A Neural Network (Deep SVDD in this case) is trained on Task $t_i$, producing optimum $\theta^*_{t_i}$. The Hessian at $\theta^*_{t_i}$ is used to obtain the top $k$ eigen directions. Next, Task $t_{i+1}$ is trained on the Neural Network that was initially trained for Task $t_i$, using the sloppy constraint mode. This results in circumventing forgetting in a task-conflicting setting.} 
     \label{fig:fig2}
 \end{figure}

Since calculating the Hessian explicitly is computationally infeasible, we compute the top‑$k$ eigenpairs using the Hessian–vector product. Our method has been compared against the baseline methods EWC, OGD, OWM, and GPM\\\\ We evaluate these two network architectures on the MNIST and CIFAR-10 datasets. These datasets were selected to assess the models' performance on both simple and complex image classification tasks.\\\\For CIFAR-10 experiments, images were converted to grayscale before training. Weight decay was not used for LeNet because the classification objective relies on learning discriminative features, and additional parameter regularization could alter the loss landscape geometry used for Hessian-based subspace extraction.\\

CIFAR-10 on LeNet was evaluated over 5 distinct 5-task combinations where each task was generated from 5 split-class pairs ([0, 1], [2, 3], [4, 5], [6, 7], [8, 9]) on CIFAR-10. MNIST on DeepSVDD was evaluated over 6 distinct 4-task combinations where each task was a digit from MNIST.  Each task is evaluated across 9 subspace dimensions ($k$ found in spectral mode factored with factors $\in \{0.1, 0.25, 0.5, 0.75, 1.0, 1.25, 1.5, 2.0, 3.0\}$) across all 3 optimization modes (sloppy, stiff, none).\\\\All experiments were executed on a CUDA-enabled NVIDIA GPU utilizing high-precision Tensor Core matmul acceleration. Implemented in Python using PyTorch, torchvision, scipy, scikit-learn, numpy, pandas, and matplotlib.\\\\ Runs use three reproducible random seeds for each sequence.

\subsection{Finding the optimal $k$}
We find the optimal $k$ for a particular dataset and model architecture using the spectral method. We compute the cumulative sum of the top $m$ (100) Hessian eigenvalues, sorted in descending order (most positive first), and select the smallest $ k$ at which this sum reaches 90\% of the total. Also, to account for $k$ being slightly greater than or less than the calculated value, we use factors to ensure all values are covered. The factors are $\{0.1, 0.25, 0.5, 0.75, 1.0, 1.25, 1.5, 2.0, 3.0\}$

\begin{table}[htbp]
\centering
\begin{tabular}{|l|l|}
\hline
\textbf{Category} & \textbf{Items} \\ \hline
\multirow{2}{*}{Models} 
    & Deep SVDD\cite{ruff2018deep}\\
    \cline{2-2}
    & LeNet\cite{lecun1998gradient} \\ \hline
\multirow{2}{*}{Datasets}
    & MNIST\cite{lecun1998gradient} \\ \cline{2-2}
    & CIFAR-10\cite{krizhevsky2009learning} \\ \hline
\end{tabular}
\caption{Models and datasets used in the experiments.}
\label{tab:models_datasets}
\end{table}

\begin{table}[htbp]
\centering
\scriptsize

\begin{tabular}{|l|c|c|}
\hline
\textbf{Parameter} & \textbf{MNIST} & \textbf{CIFAR-10} \\ 
\hline

Model & Deep SVDD Network & Multi-head LeNet \\ \hline
Input & $28\times28$ Gray & $32\times32$ Gray \\ \hline
Task Setup & Single-class & 2-class Incremental \\ \hline
Feature Dim. & 4 & 84 \\ \hline
Loss & SVDD Loss & Binary Cross-Entropy \\ \hline

Optimizer $t_0$ & Adam & Adam \\ \hline
LR $t_0$ & $10^{-3}$ & $10^{-3}$ \\ \hline
Weight Decay $t_0$ & $10^{-5}$ & None \\ \hline
Epochs $t_0$ & 10 & 10 \\ \hline

Optimizer $t_{i+1}$ & SGD & SGD \\ \hline
LR $t_{i+1}$ & $10^{-2}$ & $5\times10^{-3}$ \\ \hline
Epochs $t_{i+1}$ & 20 & 10 \\ \hline

Batch Size & 1024/1000 & 512/1000 \\ \hline

Hessian Eigenvectors & 100 & 100 \\ \hline
EWC $\lambda$ & 0.1--1000 & 1--10000 \\ \hline
GPM Threshold & 0.97 & 0.97 \\ \hline
GPM Fraction & 0.9 & 0.9 \\ \hline
OWM Alpha & $10^{-3}$ & $10^{-2}$ \\ \hline

\end{tabular}
\caption{Summary of experimental settings. $t_0$ refers to the first task and $t_{i+1}$ refers to all tasks from $t_1$ to $t_n$}
\label{tab:experiment_summary}
\end{table}

    \begin{figure}[htbp]
     \centering
     \includegraphics[width=0.75\linewidth]{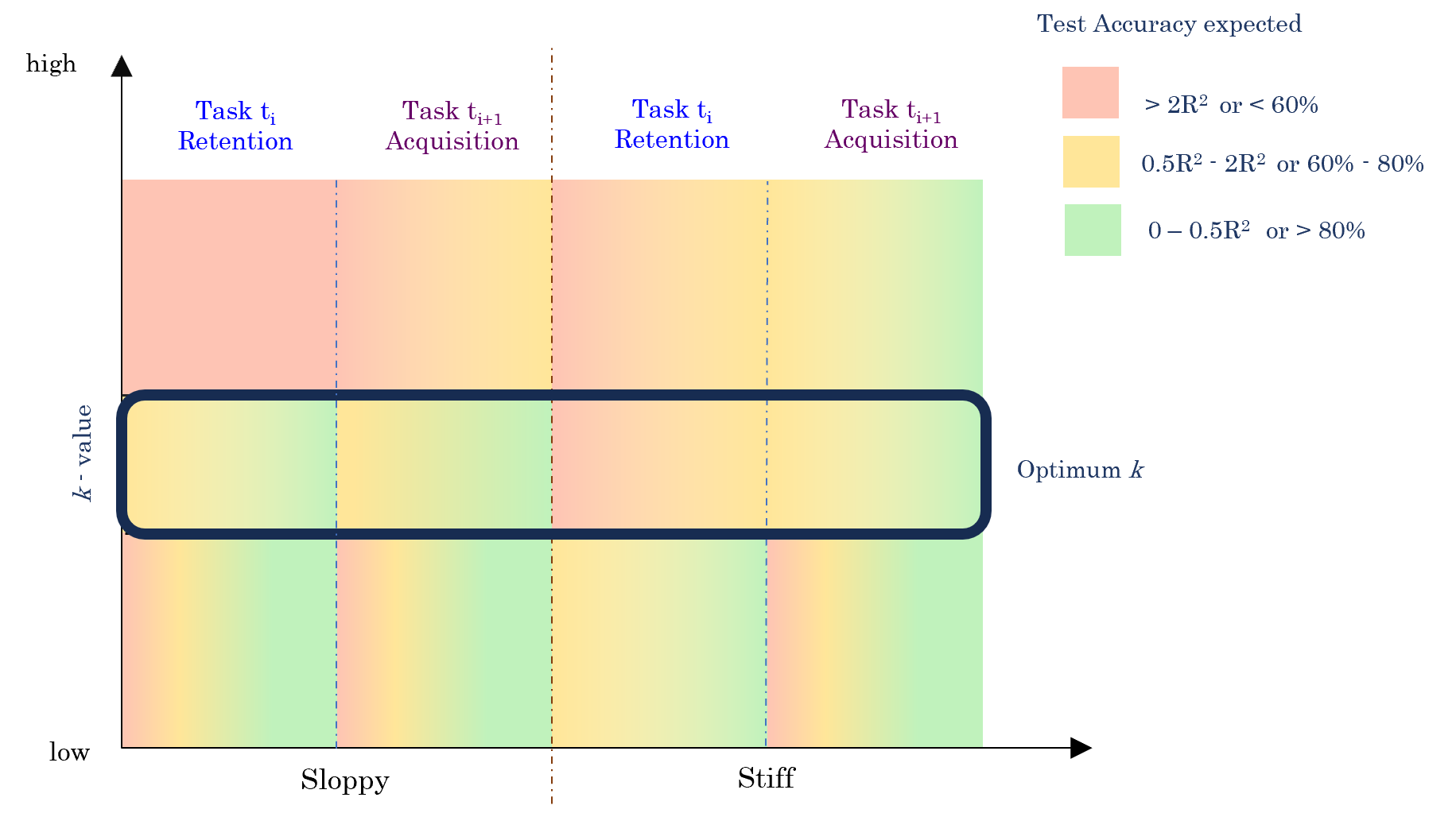}
     \caption{\textbf{Learning and Retention behavior at different values of $\mathbf{k}$}: In this figure, the vertical axis represents $k$, the number of top eigendirections selected, ranging from low to high values (relative to the network). The horizontal axis is divided into two sections to represent two constraint modes: sloppy and stiff. Each of these sections is further divided to represent how well the network performs in two aspects: the retention of Task ${t_{i}}$ and the acquisition of Task ${t_{i+1}}$. The highlighted section demonstrates the behavior of sloppy and stiff constraint modes at an optimal value of $k$} 
     \label{fig:figcc}
 \end{figure}

\section{Results}
\begin{figure}[htbp]
\centering     
\includegraphics[width=0.90\linewidth]{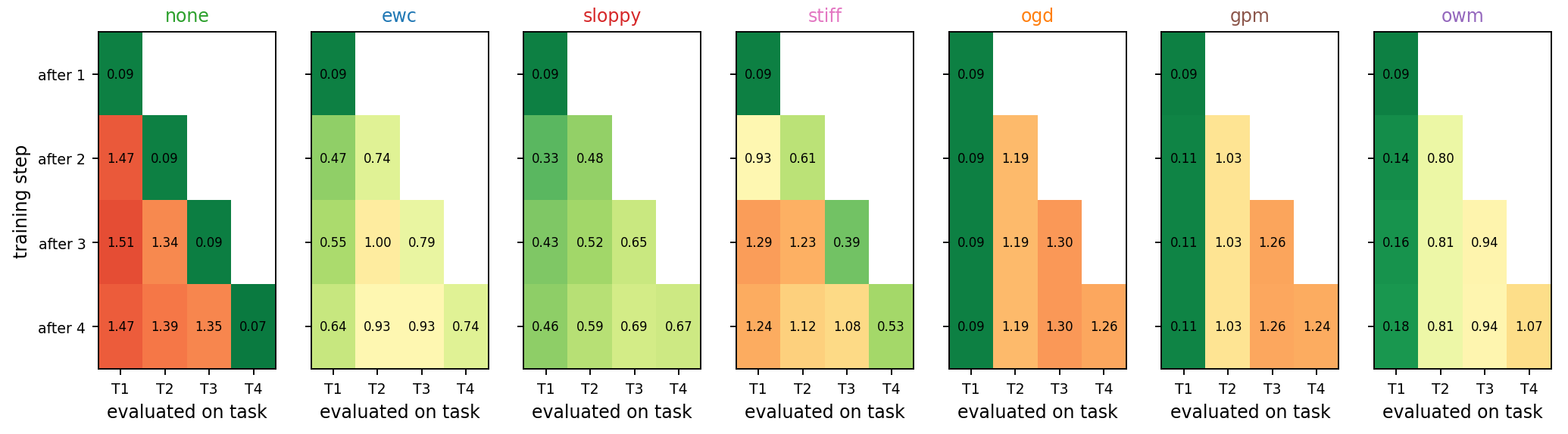}
\vspace{0.3cm}

\includegraphics[width=0.90\linewidth]{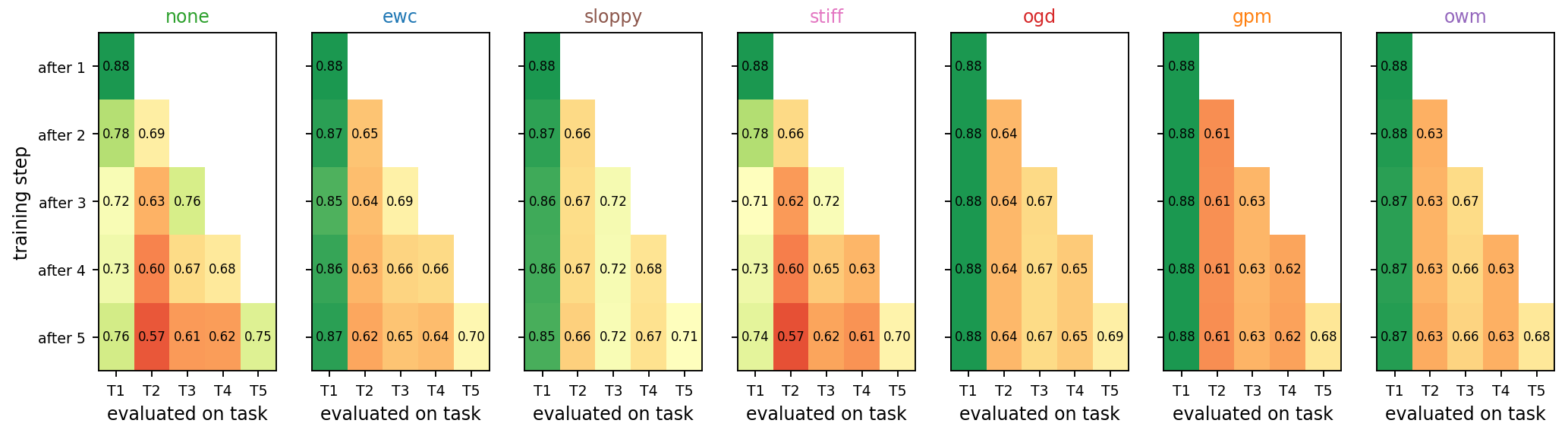}
\caption{\textbf{Performance across tasks.} (Top), MNIST on DeepSVDD, performance is given by the distance from the center, lower the better. (Bottom), CIFAR-10 on LeNet; performance is the classification accuracy, higher is better.\\}
     \label{fig:figmatrix}
 \end{figure}

\footnote{Code Repository and Full Results: \url{https://github.com/Haripriya-Narahari/Sloppiness_CF.git}}
Results against baselines EWC, OGD, OWM, GPM, and naive fine-tuning reveal that our method performs better, especially when retaining the latest task (one before the current task) compared to the other methods. It does particularly well on hard tasks that other methods fail to retain.\\\\
Figure \ref{fig:figmatrix} (top) shows the retention distance for each of the 4 tasks after training each tasks. This is averaged across 6 task sequences across 3 seeds for for MNIST on DeepSVDD. Figure \ref{fig:figmatrix} (bottom) shows the retention accuracy for 5 tasks after training each tasks. This is again averaged across 3 seeds and 5 different sequences for each method for CIFAR-10 on LeNet. The 6 MNIST task sequences, each comprising 4 tasks, and 5 CIFAR-10 task sequences, each comprising 5 tasks, are given in Table \ref{tab:digit-combinations} and \ref{tab:task_sequences}. \\ 
\begin{table}[htbp]
    \centering
    
        \begin{tabular}{|c||c|c|c|c|}
        \hline
        \textbf{Sequence \#} & \multicolumn{4}{|c|}{\textbf{Digit Combination}} \\
        \hline
    Sequence 0 & digit 1 & digit 2 & digit 3 & digit 4 \\ 
    Sequence 1 & digit 5 & digit 6 & digit 7 & digit 8 \\ 
    Sequence 2 & digit 0 & digit 2 & digit 4 & digit 6 \\ 
    Sequence 3 & digit 1 & digit 3 & digit 5 & digit 7 \\ 
    Sequence 4 & digit 4 & digit 5 & digit 6 & digit 7 \\  
    Sequence 5 & digit 2 & digit 3 & digit 4 & digit 5 \\  \hline
\end{tabular}
    \caption{MNIST Digit Combinations}
    \label{tab:digit-combinations}
\end{table}

\begin{table}[htbp]
\centering
\resizebox{\textwidth}{!}{%
\begin{tabular}{c|ccccc}
\hline
\textbf{Sequence} & \textbf{$t_0$} & \textbf{$t_1$} & \textbf{$t_2$} & \textbf{$t_3$} & \textbf{$t_4$} \\
\hline
Sequence 0 & airplane, automobile & bird, cat & deer, dog & frog, horse & ship, truck \\
Sequence 1 & ship, truck & frog, horse & deer, dog & bird, cat & airplane, automobile \\
Sequence 2 & airplane, automobile & deer, dog & ship, truck & bird, cat & frog, horse \\
Sequence 3 & frog, horse &  bird, cat & ship, truck & airplane, automobile & deer, dog \\
Sequence 4 & deer, dog & ship, truck & airplane, automobile & frog, horse & bird, cat \\
\hline
\end{tabular}%
}
\caption{CIFAR-10 Task sequences}
\label{tab:task_sequences}
\end{table}
    
\begin{figure} [htbp]
\centering     
\includegraphics[width=0.55\linewidth]{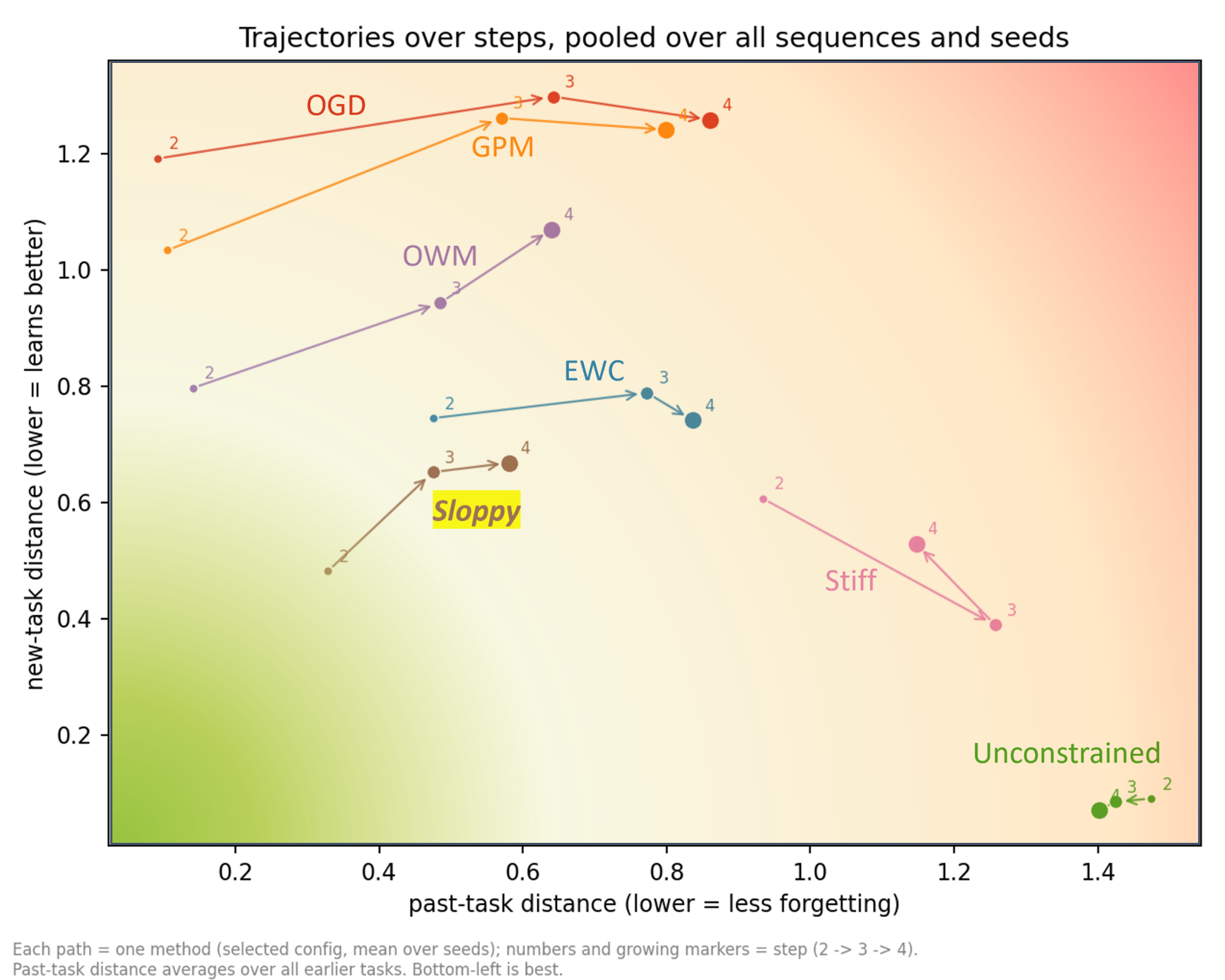}
     \caption{\textbf{Aggregated Results for MNIST on DeepSVDD for 6 sequences of four tasks.}  The x-axis represents past task distance from the center and the y-axis the new task distance. The green region is the most ideal performance (lower distance)}
    
     \label{fig:figMNIST}
 \end{figure}

\begin{figure} [htbp]
\centering     
\includegraphics[width=0.55\linewidth]{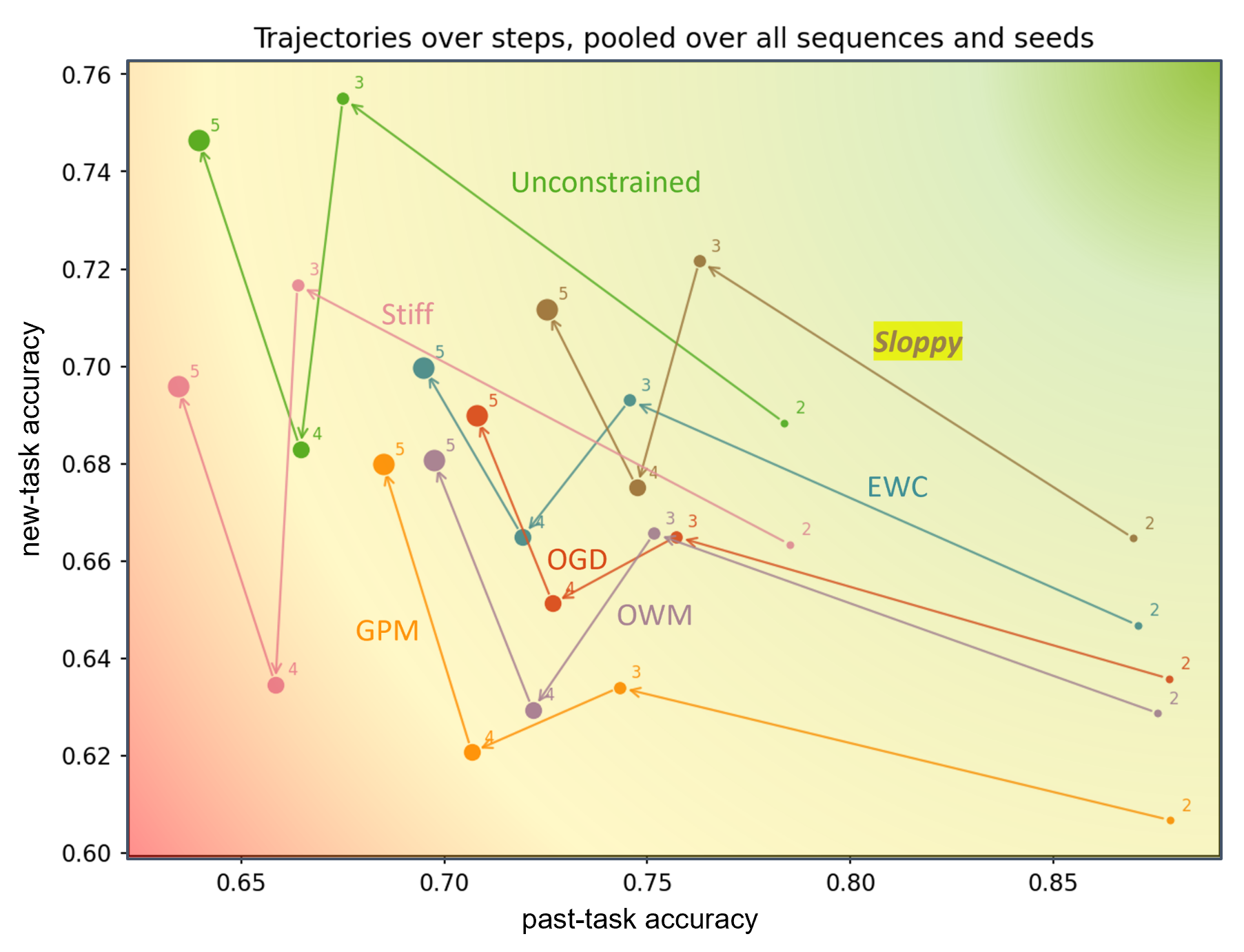}
     \caption{\textbf{Aggregated Results for CIFAR-10 on LeNet for 5 sequences of five tasks-pairs.} The x-axis represents past task accuracy and the y-axis the new task accuracy. The green region is the most ideal performance (higher accuracy)
    }

     \label{fig:figCIFAR}
 \end{figure}

Table \ref{tab:main_results} shows that our method performs better than baselines EWC, OGD, OWM, GPM, and naive fine-tuning.
In Figures \ref{fig:figCIFAR} and \ref{fig:figMNIST}, we plot the average retention of $n-1$ tasks vs. the learning performance of the $n^{th}$ task pair across different benchmarks and averaged over three seeds over the best configurations ($k$ and $\lambda$). The green region in the graphs indicates the optimal performance where both average retention of previous tasks and learning of new tasks have optimal performance.
From schematic Figure \ref{fig:figcc}, we observe that the stiff mode exhibits poor retention even with the optimal $k$ value and good retention with $<k$ value. This means that when some important directions were not captured in the stiff subspace with $<k$, the ones not captured were protected; hence the unaffected performance. When the optimum $k$ is chosen, all important (Stiff directions) are captured, leading to all these directions being modified when projecting in the stiff subspace, causing forgetting. This shows that parameter updates in stiff directions are indeed responsible for catastrophic forgetting. To ensure no movement occurs in stiff directions in the sloppy mode and in sloppy directions in the stiff mode, we calculate the angle between the update direction and the stiff direction. An angle of $0^\circ$ means movement in the stiff direction, and $90^\circ$ means movement orthogonal to the stiff direction, i.e., the sloppy direction, and we observe that the projection preserves the expected angle in both cases.\\ 
\begin{table}[htbp]
\centering
\scriptsize
\resizebox{\textwidth}{!}{%
\begin{tabular}{lcccc}
\toprule
\textbf{Method} & \multicolumn{2}{c}{\textbf{MNIST on DeepSVDD}} & \multicolumn{2}{c}{\textbf{CIFAR on LeNet}} \\
\cmidrule(lr){2-3} \cmidrule(lr){4-5}
 & \textbf{Retention loss (dist) for 3 tasks $t_1$, $t_2$, $t_3$}  & \textbf{Learning loss (dist) for new task $t_{4}$} & \textbf{Retention accuracy(\%) for 4 tasks $t_0$, $t_1$, $t_2$, $t_3$ } & \textbf{Learning accuracy(\%) for task $t_{4}$} \\
\midrule
Naive Fine-Tuning & $1.406 \pm 0.019$ & $0.0753 \pm 0.007$ & $64.5 \pm 0.37$ & $82.05 \pm 0.31$ \\
EWC (Kirkpatrick et al. 2017) & $0.8071 \pm 0.045$ & $0.5636 \pm 0.0733$ & $69.41 \pm 0.73$ & $75.06 \pm 2.29$ \\
OGD (Farajtabar et al. 2020) & $0.8013 \pm 0.0374$ & $1.0353 \pm 0.1813$ & $70.38 \pm 0.51$ & $76.11 \pm 1.710$ \\
GPM (Saha, et al. 2021) &$0.6982 \pm 0.029$ & $1.0977 \pm 0.06$ & $69.23 \pm 0.856$ & $75.6 \pm 1.99$ \\
OWM (Zeng et al. 2019) &$0.5438 \pm 0.054$ & $0.9677 \pm 0.054$ & $69.84 \pm 0.684$ & $76.25 \pm 2.042$ \\
Stiff Mode & $1.2129 \pm 0.2798$ & $0.3945 \pm 0.3994$ & $60.59 \pm 5.99$ & $72.78 \pm 11.90$ \\
\underline{\textbf{Sloppy Mode (Ours)}} & $\mathbf{0.5757 \pm 0.0456 }$ & $\mathbf{0.5386 \pm 0.1263}$ & $\mathbf{71.22 \pm 0.07}$ & $\mathbf{78.61 \pm 1.508}$ \\
\bottomrule
\end{tabular}%
}
\caption{\textbf{Comparison of sequential learning performance on MNIST and CIFAR across all benchmarks for a single sequence (Seq 0)}. The results have been aggregated across 3 seeds for a single 4-sequence task on DeepSVDD and for a single 5-sequence task on LeNet: 4 MNIST tasks (digit $1$, digit $2$, digit $3$, digit $4$), and 5 CIFAR tasks, with each task being a pair of images - $
\left\{ \{ \text{image}_i, \text{image}_{i+1} \} \mid i = 0, 2, \ldots, n-2 \right\}$
For EWC, Sloppy and Stiff $\lambda_{\text{best}}$, $k_{\text{sloppy-best}}$, and $k_{\text{stiff-best}}$ that produced the best result for each task were picked and aggregated.\\ }
\label{tab:main_results}
\end{table}


\begin{figure}[htbp]
    \centering

    \includegraphics[width=0.95\linewidth]{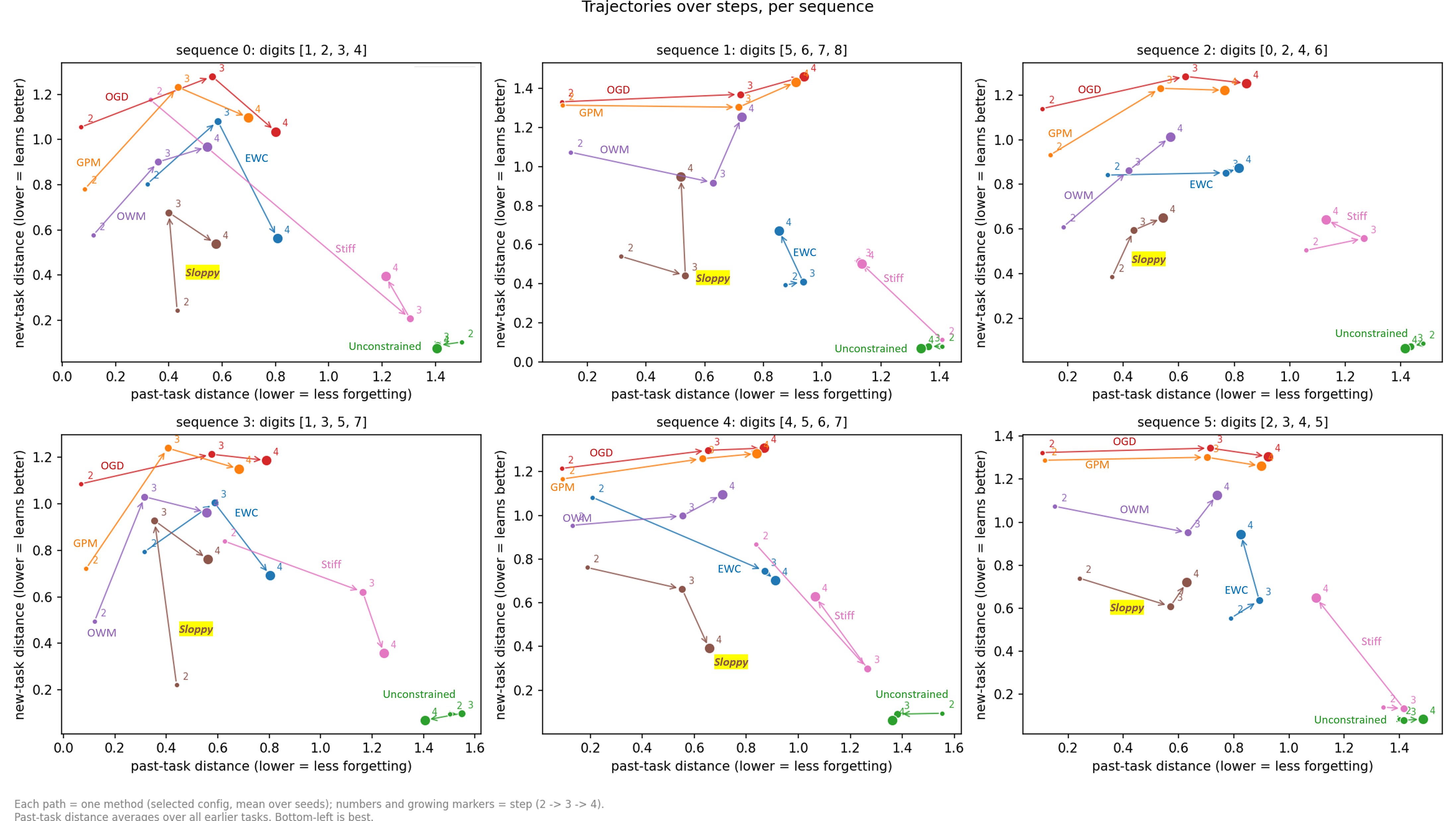}

    \vspace{0.3cm}

    \includegraphics[width=0.95\linewidth]{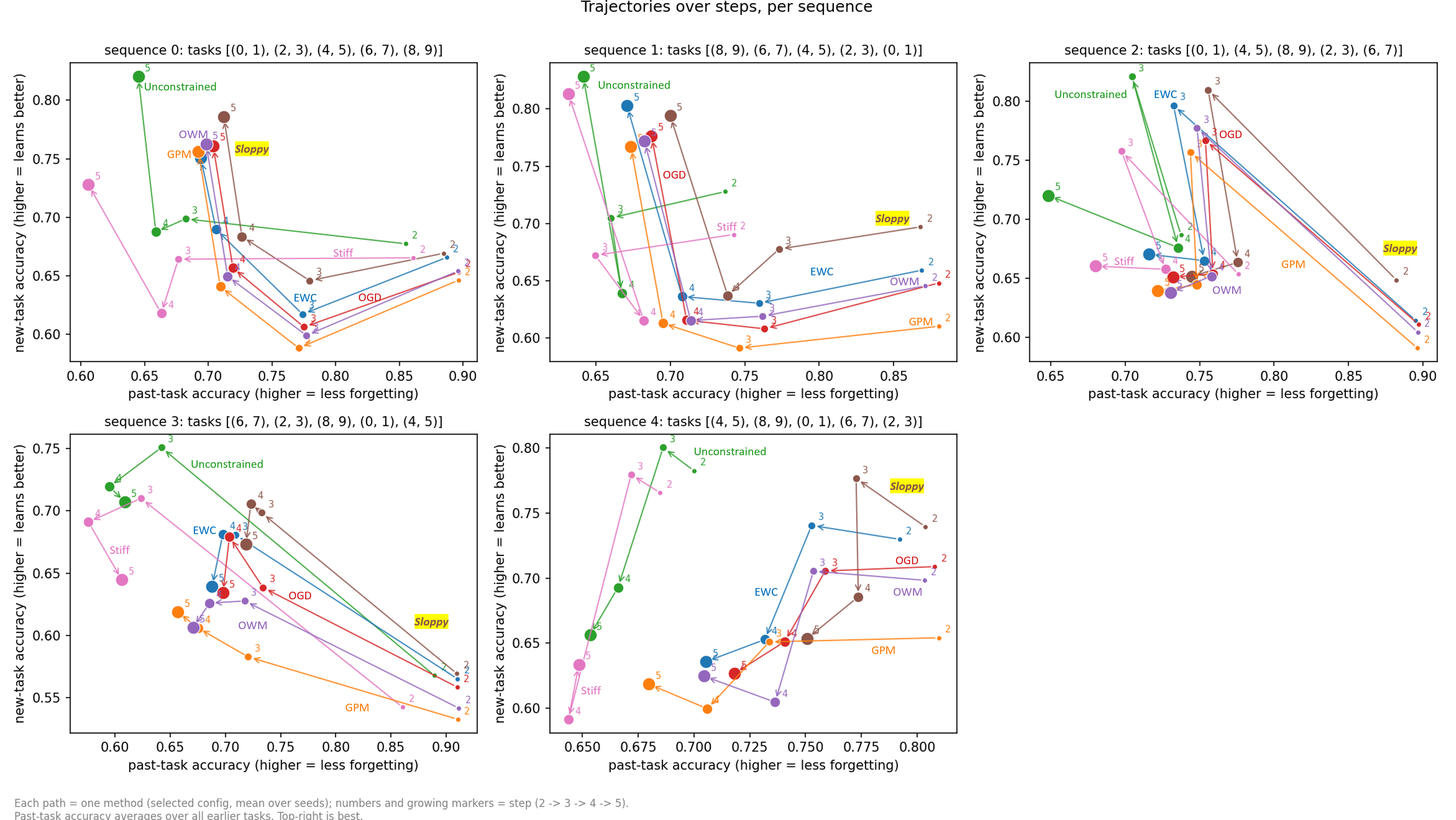}

    \caption{Average past task retention performance vs new task learning performance plotted for all methods after every training step, averaged across three seeds. (Top) MNIST on DeepSVDD across six sequences of four tasks. (Bottom) CIFAR on LeNet across five sequences of five tasks.}
    \label{fig:trajectories_MNIST_CIFAR}
\end{figure}

\begin{table}[htbp]
\centering
\begin{tabular}
{|p{0.18\textwidth}|p{0.18\textwidth}|p{0.18\textwidth}|p{0.18\textwidth}|}
\hline
\multicolumn{2}{|c|}{} & \multicolumn{1}{c|}{Sloppy}  & \multicolumn{1}{c|}{ Stiff} \\
\hline
\multirow{2}{*}{$k$ value} 
& Low 
& Few important stiff directions are included here 
& All important directions are not captured \\
\cline{2-4}
& High 
& Important moderate curvature directions are lost for task retention 
& All high curvature and moderate curvature directions are captured \\
\hline
\end{tabular}

\caption{Effect of $k$ value on sloppy and stiff directions}
\end{table}

Studies \cite{wu2020dissecting, gur2018gradient,sagun2017empirical} have shown that only the top $k$ eigen directions ($k \ll n$), where $n$ is the total number of learnable parameters in the neural network, dominate the Hessian of the NN and capture most of the curvature of the loss landscape. \cite{ghorbani2019investigation} also shows that a few large-valued eigenvalues and the majority close to zero dominate the Hessian spectrum. These directions correspond to the so-called \emph{stiff} directions of the network. 

\section{Future Work}
There is an observation of the general stability-plasticity dilemma \cite{mermillod2013stability}, where better retention of old tasks can lead to a slight drop in plasticity. The choice of $k$ at which a good balance of old task retention and new task learning can be achieved is dependent on the dataset and the number of parameters in the NN.\\\\Computing the Hessian-vector product to assess the curvature and projecting the network into stiff and sloppy subspaces is computationally intensive, which calls for a more efficient process. Additionally, the above assumptions made, which may not hold in every real-world case, also play a major role in mitigating catastrophic forgetting. Future work would therefore focus on reducing the computational cost while developing methods that are more robust to violations of these assumptions.
\section{Acknowledgement}
HPN acknowledges the Post-Baccalaureate fellowship from the Wadhwani School of Data Science and Artificial Intelligence. 

\section{AI Use Statement}
In this work, we used generative AI tools for assistance with code development and plot generation. We did not use generative AI
tools for research ideation or experimental design. Additionally, we used generative AI tools for assistance in grammar and polishing the manuscript. We have reviewed all AI-assisted work and take responsibility for the final content of this work, including text, claims, or artifacts produced with the aid of generative AI.

\bibliography{ref}
\bibliographystyle{unsrt}

\end{document}